\documentclass[british]{elsarticle}
\usepackage[T1]{fontenc}
\usepackage[latin9]{inputenc}
\usepackage{geometry}
\usepackage{babel}
\usepackage{url}
\usepackage{graphicx}
\usepackage[unicode=true,
 bookmarks=true,bookmarksnumbered=false,bookmarksopen=false,
 breaklinks=false,pdfborder={0 0 0},backref=false,colorlinks=false]
 {hyperref}

\makeatletter

\providecommand{\tabularnewline}{\\}
\newcommand{\lyxdot}{.}

\makeatother

\begin{document}

\title{Glasgow's Stereo Image Database of Garments}

\author{Gerardo Aragon-Camarasa, Susanne B. Oehler, Yuan Liu, \\
Sun Li, Paul Cockshott and J. Paul Siebert}

\address{Sir Alwyn Williams Building, Lilybank Gardens, Glasgow, G12 8QQ,
Scotland}
\begin{abstract}
To provide insight into cloth perception and manipulation with an
active binocular robotic vision system, we compiled a database of
80 stereo-pair colour images with corresponding horizontal and vertical
disparity maps and mask annotations, for 3D garment point cloud rendering
has been created and released. The stereo-image garment database is
part of research conducted under the EU-FP7 \textbf{Clo}thes \textbf{Pe}rception
and \textbf{Ma}nipulation (CloPeMa) project and belongs to a wider
database collection released through CloPeMa%
\thanks{www.clopema.eu%
}. This database is based on 16 different off-the-shelve garments.
Each garment has been imaged in five different pose configurations
on the project's binocular robot head. A full copy of the database
is made available for scientific research only at \url{https://sites.google.com/site/ugstereodatabase/}. 
\end{abstract}
\maketitle

\section{Introduction\label{sec:Introduction}}

The CloPeMa project is advancing the state of the art in clothes perception
and manipulation by delivering a novel robotic system that accomplishes
automatic sorting and folding of a laundry heap. To this end, CloPeMa
is using a prototype robot composed mainly of off-the-shelf components
comprising an active binocular vision robot head. This active binocular
robot head, which is inspired by the system developed in \citet{Aragon-Camarasa2010},
has been designed by the Computer Vision and Graphics Group (CV\&G)
at the University of Glasgow. This robotic head, as created for CloPeMa,
is not only able to provide high-resolution intensity images of the
robot's workspace, as required for intensity based computer vision
algorithms, but is capable of automatic vergence and gaze control,
hand eye calibration and 2.5D reconstruction of areas-of-interest.
Data captured by this robotic head can be used in a wide variety of
applications such garment spreading and flattening\citet{Sun2013},
automatic visual inspection and exploration of cluttered scenes\citet{Aragon-Camarasa2010},
selection of better grasping points or more detailed feature extraction
and classification. In order to provide a first insight into the type
and quality of data produced by the binocular robot head in the CloPeMa
robot system, we have compiled and released a freely available database
of stereo-pair images of garments. The aim of this dataset is to serve
as a benchmark tool for algorithms for recognition, segmentation and
various range image properties of non-rigid objects. For instance,
it will be used to improve the Vector-Pascal \citet{Cockshott2012}
Glasgow parallel stereo matcher and its GPU implementations. This
dataset is the first high resolution stereo-pair garment image dataset
that is released for research purposes and potentially allows for
a variety of research applications. Therefore, the Glasgow's Stereo
Image Database of Garments can be downloaded from:

\begin{center}
\url{https://sites.google.com/site/ugstereodatabase/}.
\par\end{center}

This database comprises images of 16 different off-the-shelve garments
selected from the official CloPeMa cloth heap, defined in \citet{Molfino2012}.
The CloPeMa heap features a wide variety of textile materials with
different texture, colour and reflectance characteristics in order
to give a realistic sample of the real world clothing variety. For
the released database, the chosen garments where imaged in five possible
pose configurations: \emph{flat on the table, folded in half, completely
folded, randomly wrinkled and hanging over the robot's arm}. These
configurations are an approximation of the most representative pose
configurations a robot may encounter while sorting and folding clothes.
Each of the selected five configurations were imaged under software-control
capture synchronisation. The database therefore yields a total of
80 stereo-pairs of garment images. For completeness, the horizontal
and vertical disparities without mask of the Glasgow's Stereo Image
Database of Garments can also be downloaded.

The 80 stereo-pairs in the database have all been processed using
the Glasgow stereo matcher \citet{Cyganek2011}, in order to compute
the horizontal and vertical disparities. A new version of the Glasgow
stereo matcher has been integrated in CloPeMa's robot system as a
ROS node within the CloPeMa robot head package collection%
\footnote{\url{http://clopema.felk.cvut.cz/redmine/projects/clopema/wiki/Technical_Stuff}%
}. Additionally, the data-set\textquoteright{}s image pairs are accompanied
by mask annotations for the left as well as the right image. The camera
calibration, which has been computed using CloPeMa's integrated OpenCV
compatible robot head calibration system, is also released as part
of the database. This enables the research community to use the database
for 3D garment point cloud projection. Specifically for this purpose,
Matlab-based reconstruction software is also distributed within this
database.

It must be noted that the above algorithms and methods have been integrated
as part of a collection of ROS nodes distributed in the official CloPeMa
package collection. Specifically, the CloPeMa active robot head system
software includes ROS nodes for directing the robot's gaze under program
control, automatic vergence, acquiring synchronised stereo-pair images,
camera and hand-eye calibration routines, stereo image processing
algorithms (including a GPU stereo matcher based on the Glasgow Stereo
Matcher) , real-time SIFT feature extraction and user interactive
interfaces for gaze control and calibration routines. The robot head
ROS packages can be downloaded from:

\begin{center}
\url{http://clopema.felk.cvut.cz/redmine/projects/clopema/wiki/Packages_instalation_%28hydro%29_}
\par\end{center}

\section{Database Acquisition}

\begin{figure}
\centering\includegraphics[width=8cm]{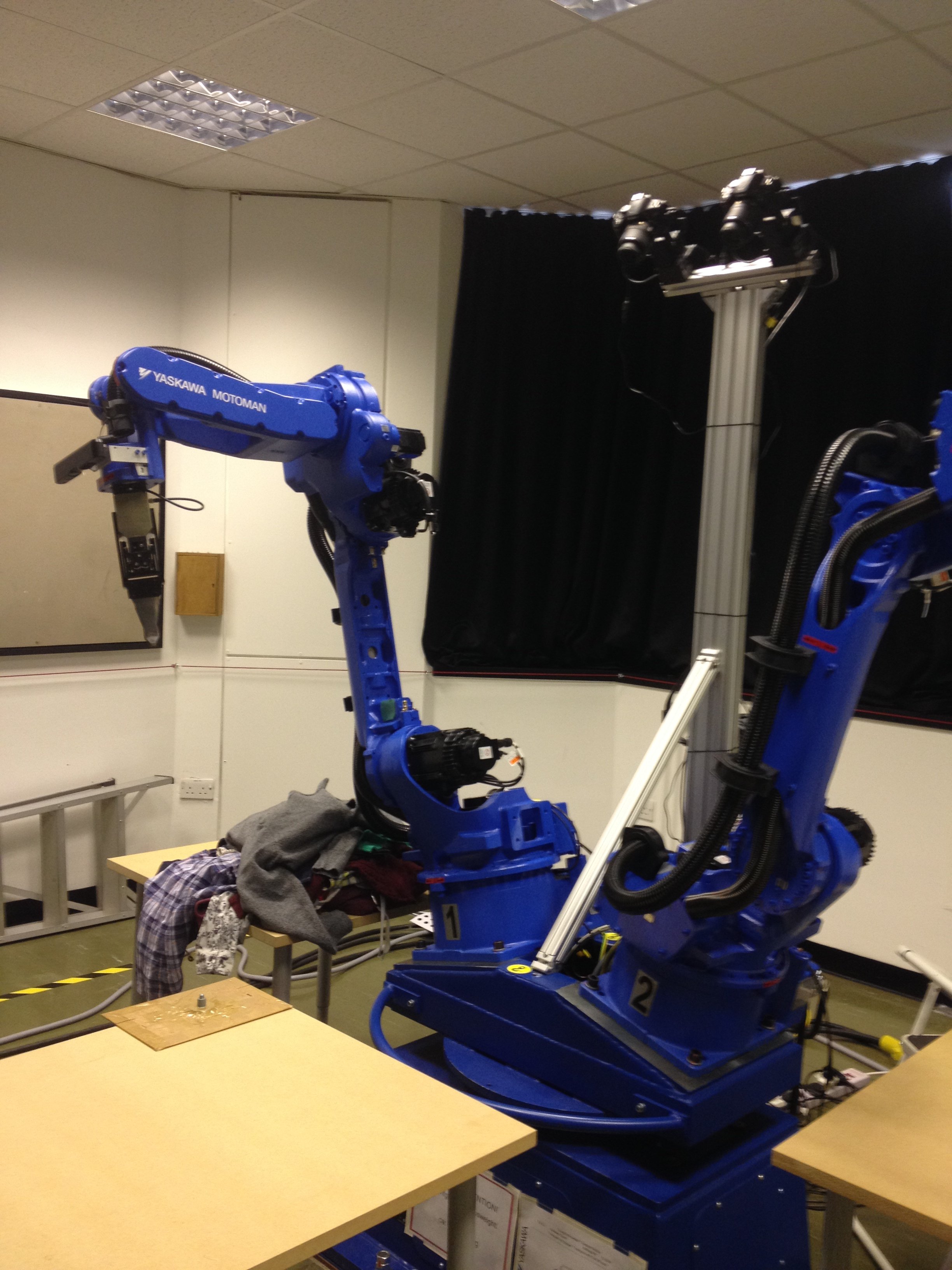}

\caption{CloPeMa test-bed at the University of Glasgow.\label{fig:CloPeMa-testbed-at}}
\end{figure}

\begin{figure}
\centering\includegraphics[width=8cm]{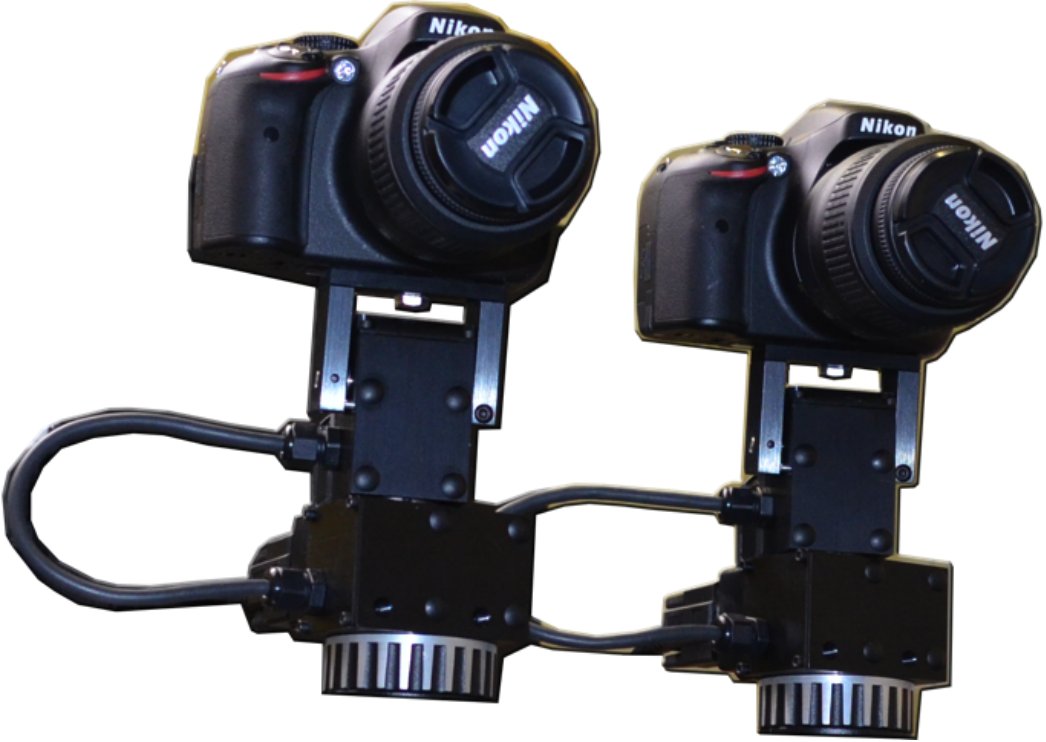}

\caption{CloPeMa robot head.\label{fig:CloPeMa-robot-head.}}
\end{figure}

The CloPeMa robotic test-bed is equipped with two Yaskawa robotic
arms mounted on a computer controllable tailor-made Yaskawa turn table,
two RGB-D sensors for wide vision mounted on the wrists of the robotic
arms, two prototype grippers designed by the University of Genoa \citet{Le2013}
and an active binocular robot head for foveated vision designed by
the University of Glasgow. Figure \ref{fig:CloPeMa-testbed-at} shows
the University of Glasgow robotic infrastructure. The database subject
of this report has been captured using the active binocular robot
head. This robot head comprises two Nikon DSLR cameras (D5100) that
are capable of capturing images at 16 Mega Pixels at different zoom
settings (manually selected, 35mm used for this database). These are
mounted on two pan and tilt units (PTU-D46) with their corresponding
controllers as depicted in Figure \ref{fig:CloPeMa-robot-head.}.
The cameras are separated by a pre-defined baseline for optimal stereo
capturing within the robot's workplace. The baseline separation between
cameras is 30 centimetres.

\begin{figure}
\centering%
\begin{tabular}{cc}
\includegraphics[width=5cm]{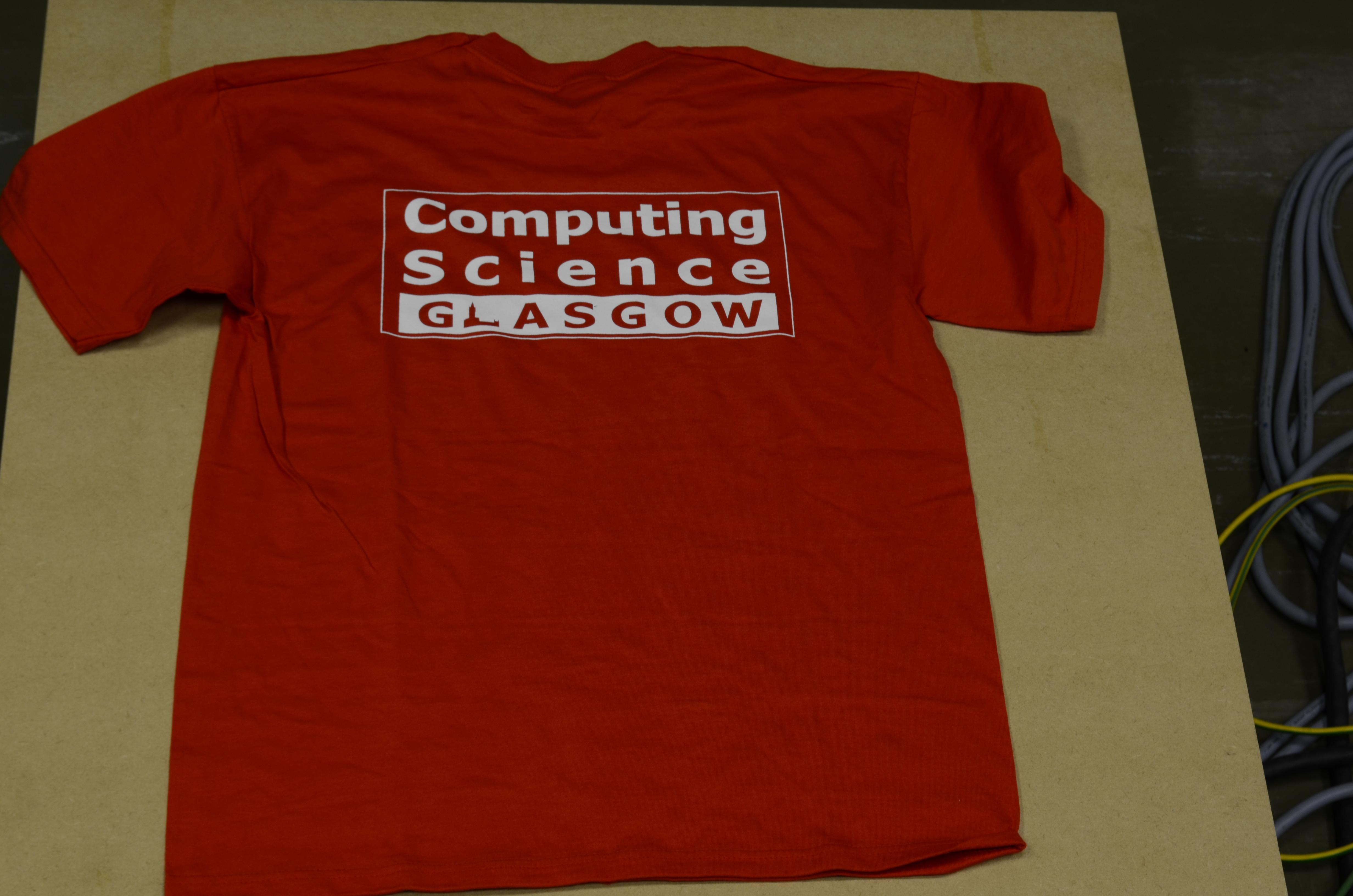} & \includegraphics[width=5cm]{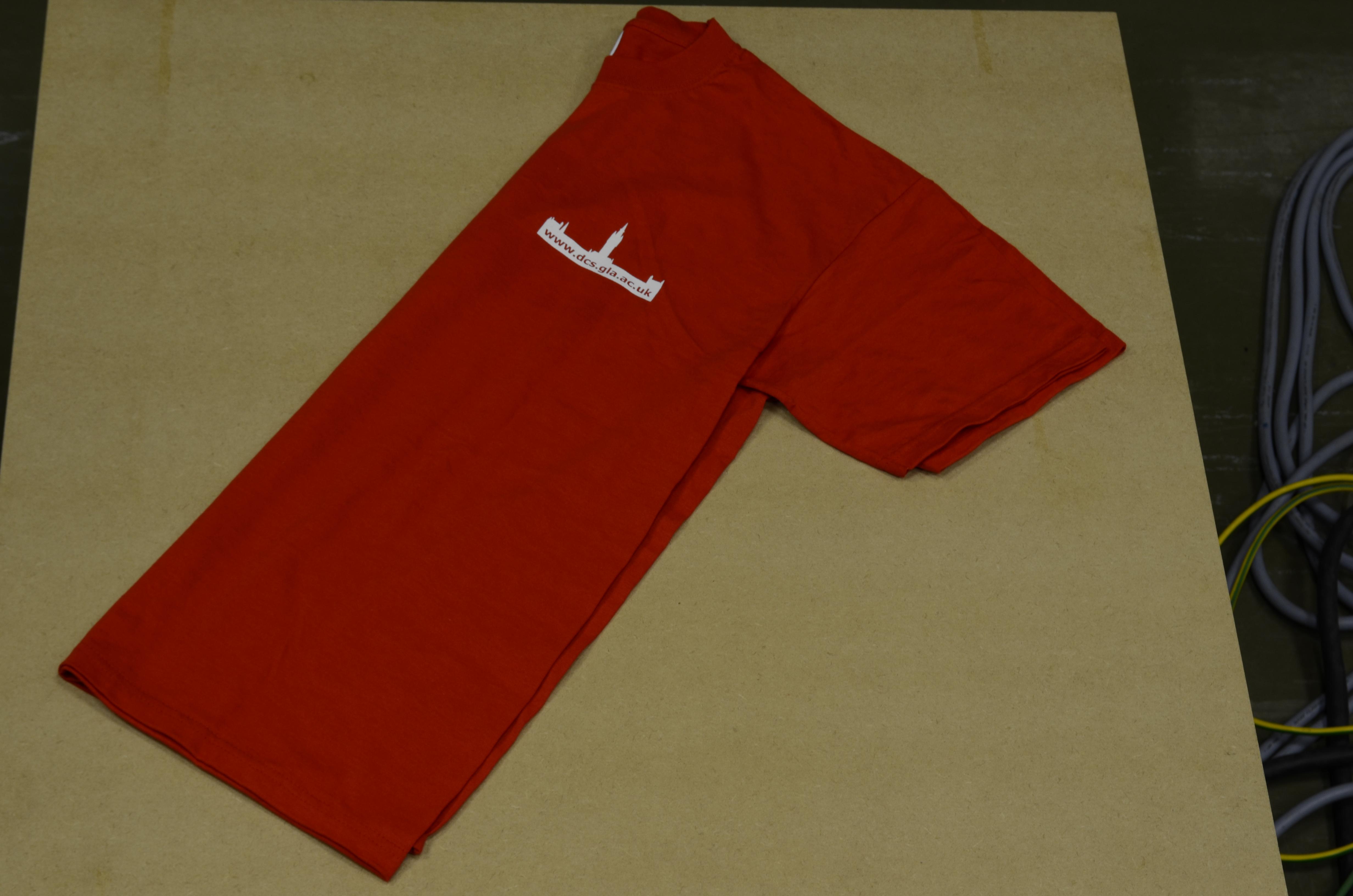}\tabularnewline
(a) Spread & (b) Half-way folded\tabularnewline
\includegraphics[width=5cm]{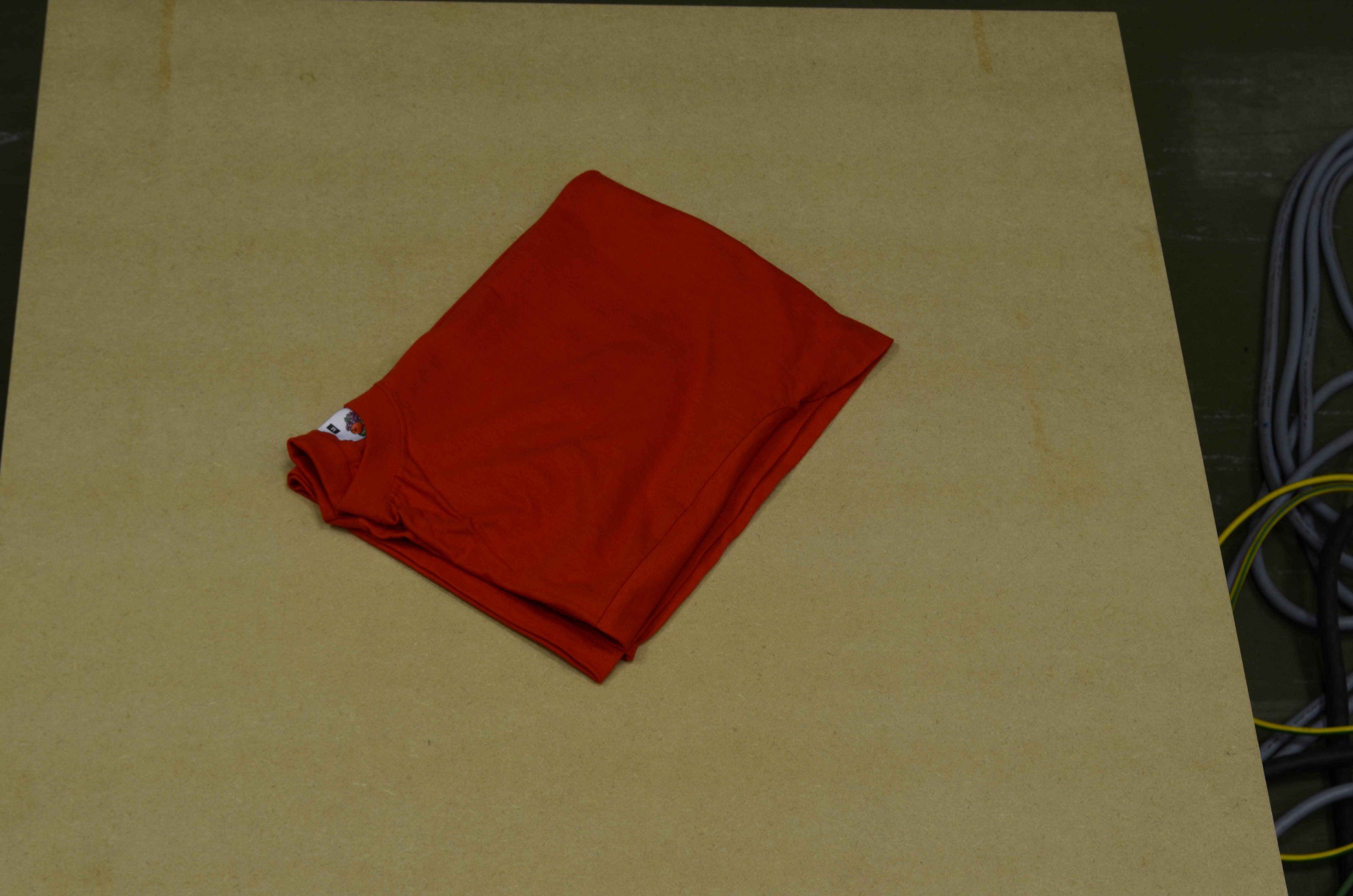} & \includegraphics[width=5cm]{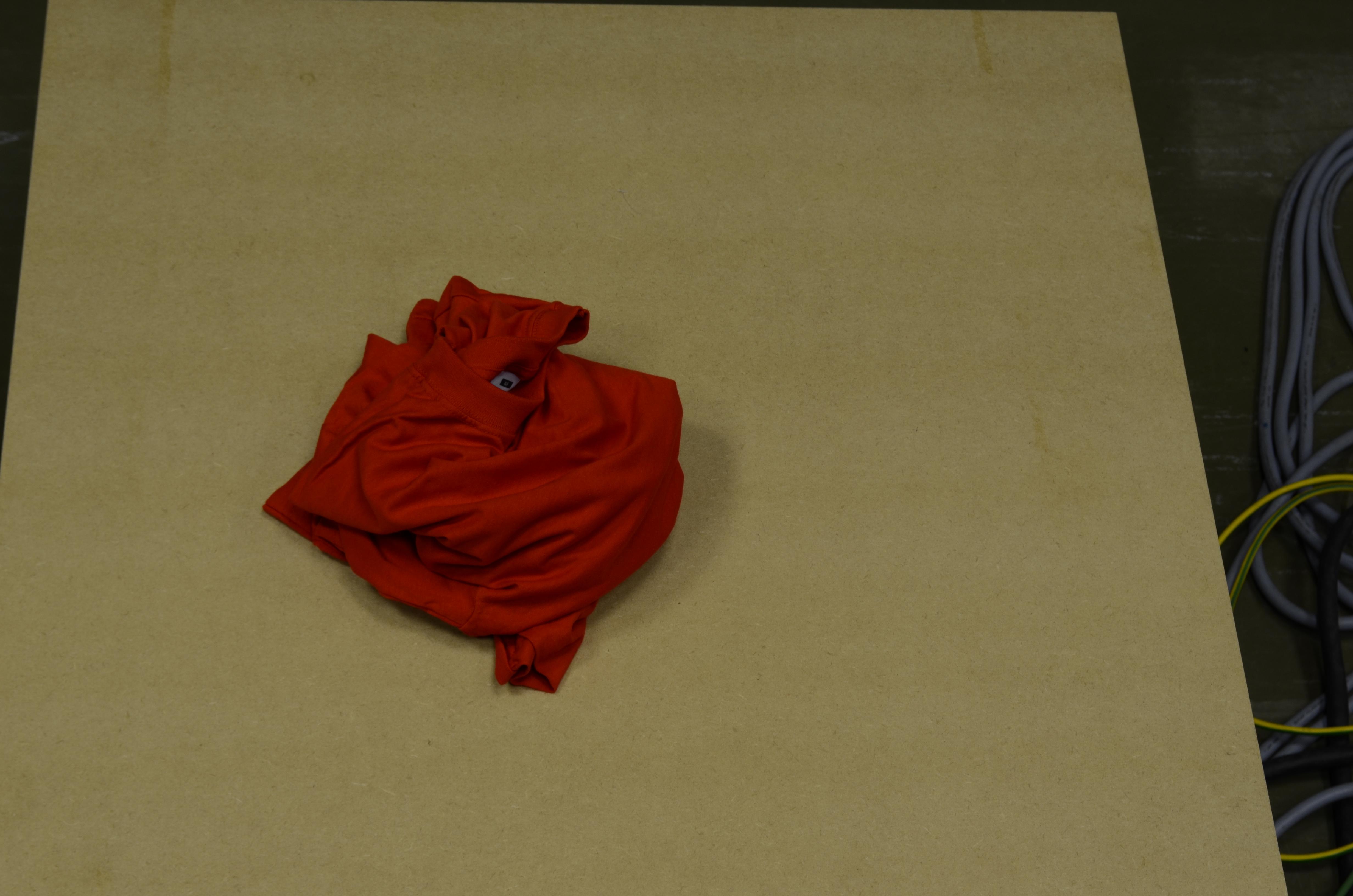}\tabularnewline
(c) Folded & (d) Wrinkled\tabularnewline
\multicolumn{2}{c}{\includegraphics[width=5cm]{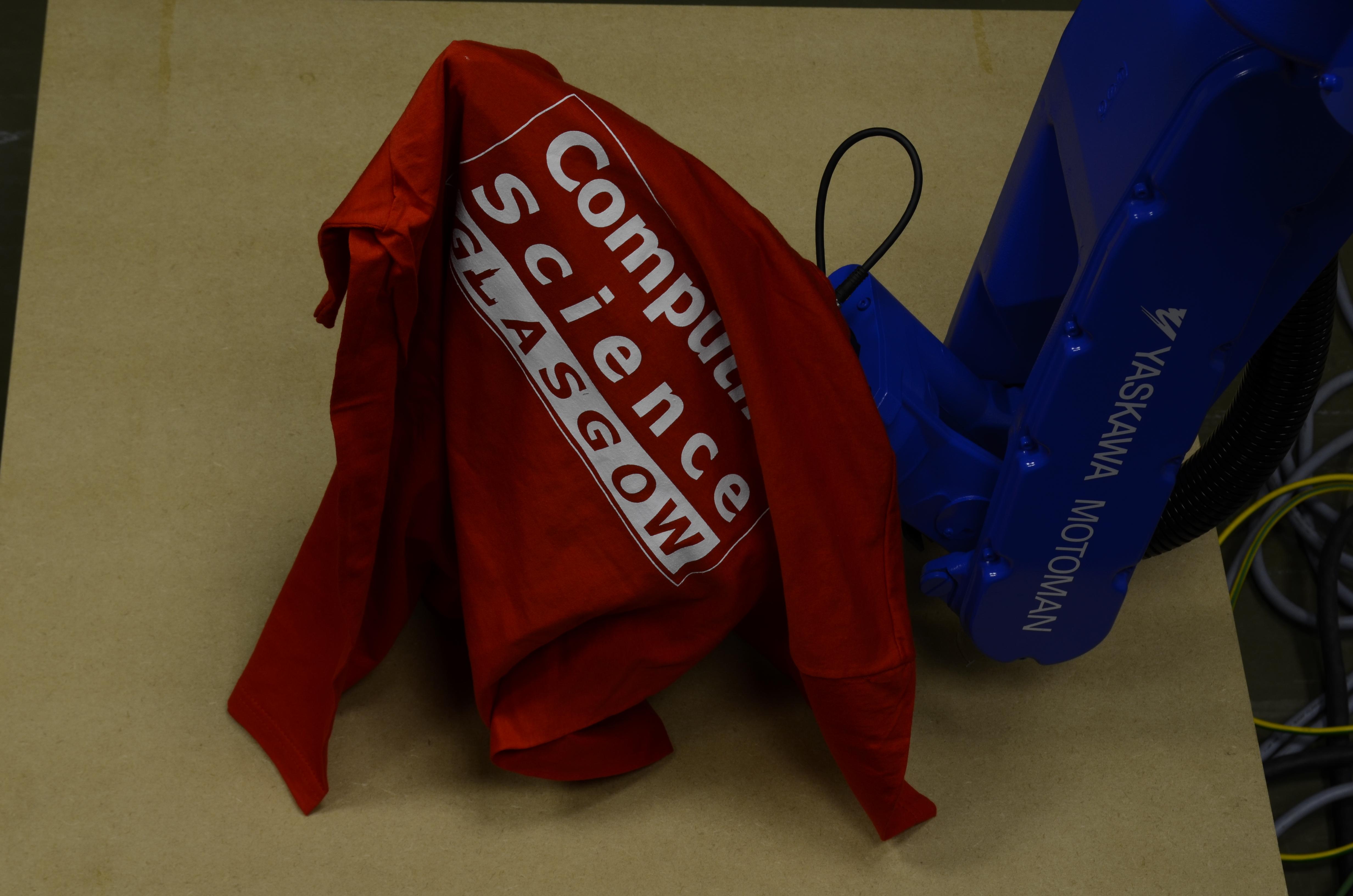}}\tabularnewline
\multicolumn{2}{c}{(e) Hanging}\tabularnewline
\end{tabular}

\caption{Garment states captured.\label{fig:Garment-states}}
\end{figure}

\begin{figure}
\centering\includegraphics[width=8cm]{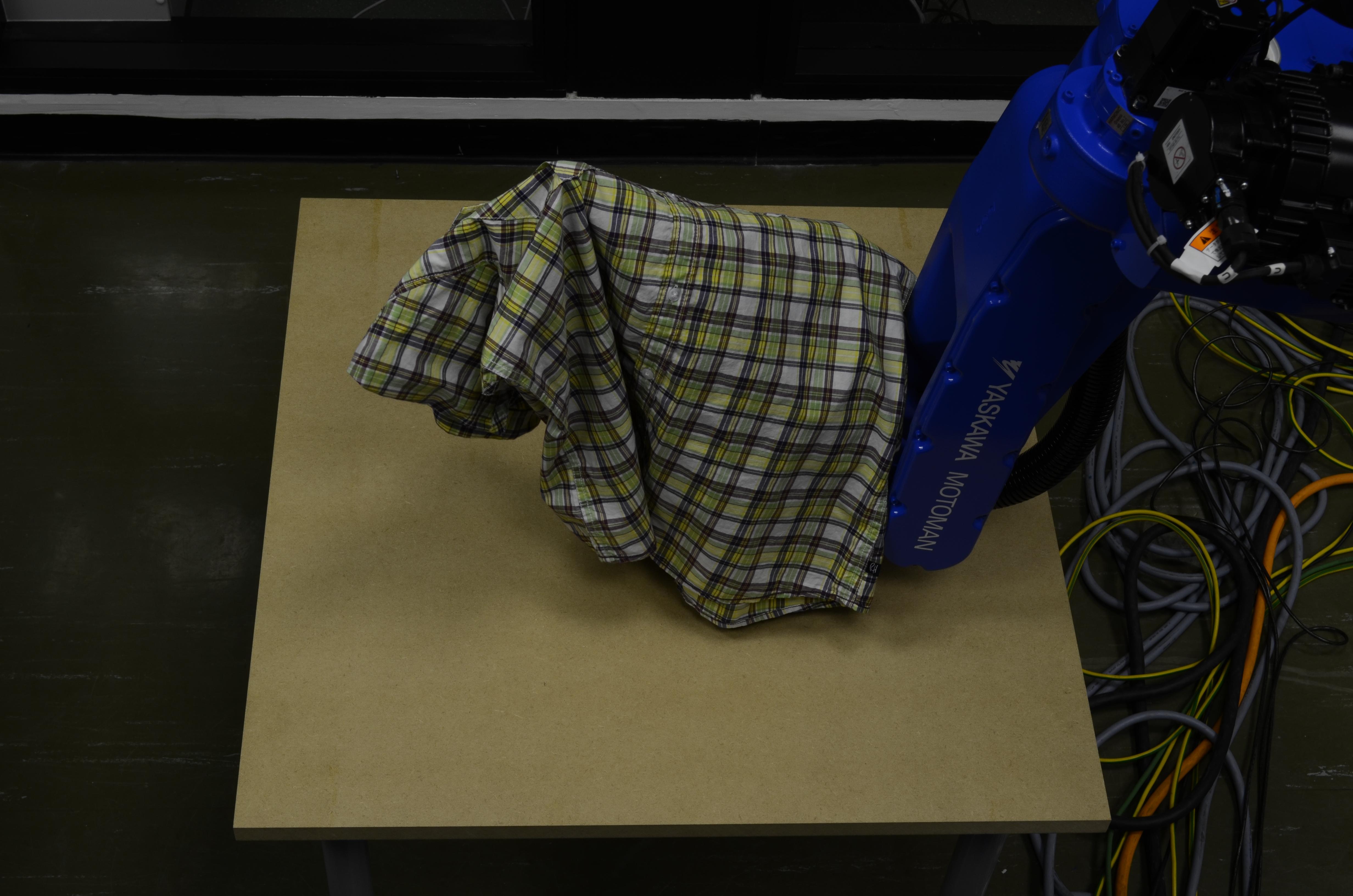}

\caption{Garment zoom setting at 35-mm. A standard Nikon 8-55mm VR lens was
used for capturing the database.\label{fig:Zoom-setting-at}}
\end{figure}

Garments were placed on a planar surface at an average distance of
1.8 meters from the binocular robot head. For each garment, five different
garment pose configurations were captured as showed in Figure \ref{fig:Garment-states}.
Figure \ref{fig:Zoom-setting-at} shows an example of the 35-mm zoom
setting. The cameras of the robotic head were converged at the centre
point in the left and right cameras prior capturing the stereo images.
For this purpose, the vergence algorithm reported in \citet{Aragon-Camarasa2010}
was used and integrated as a ROS node as described in Section \ref{sec:Introduction}.

For each image, a manually segmented mask of the same image resolution
has been provided for annotation purposes. In the database creation,
the mask was applied as part of the vertical and horizontal disparities
computation using the Glasgow stereo matcher \citet{Cyganek2011}.
\emph{Gimp 2.8}%
\footnote{\url{http://www.gimp.org/}%
} was used to segment and annotate the stereo-pair images.

\begin{figure}
\centering%
\begin{tabular}{cc}
\includegraphics[width=7cm]{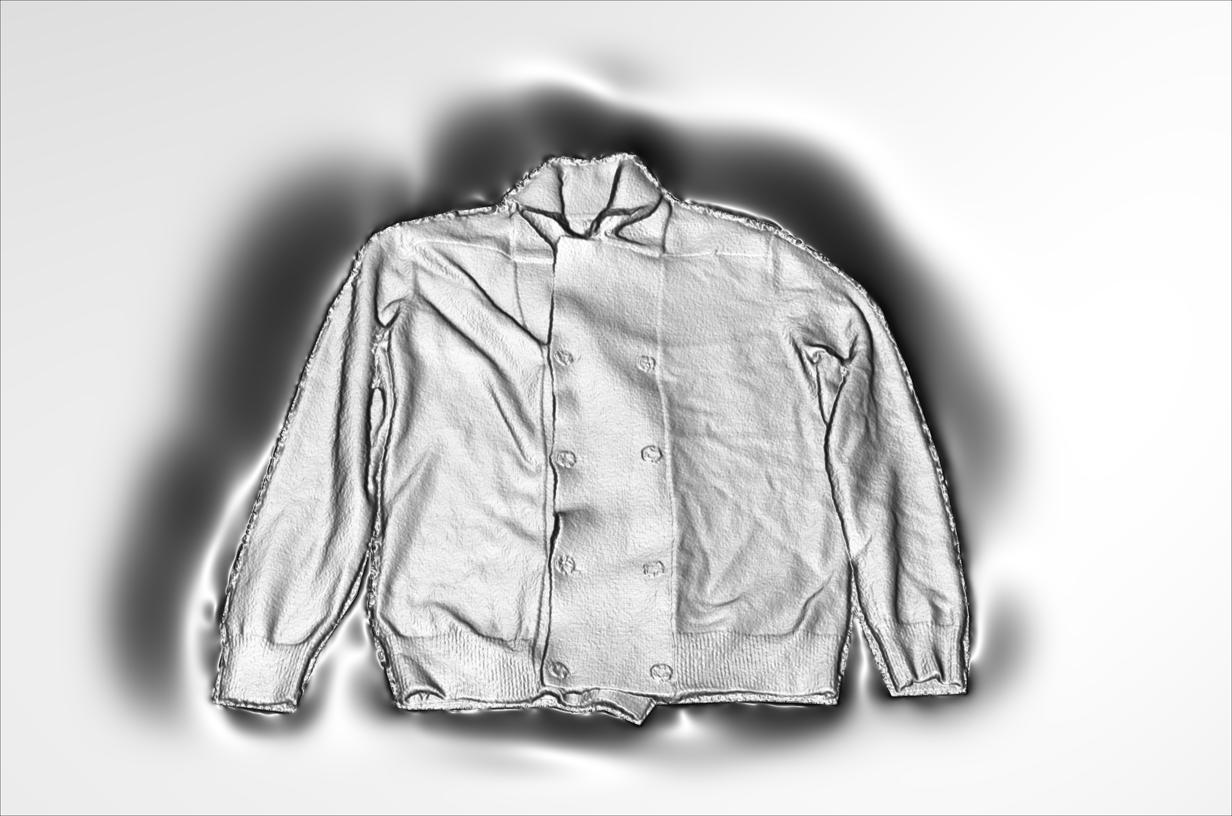} & \includegraphics[width=7cm]{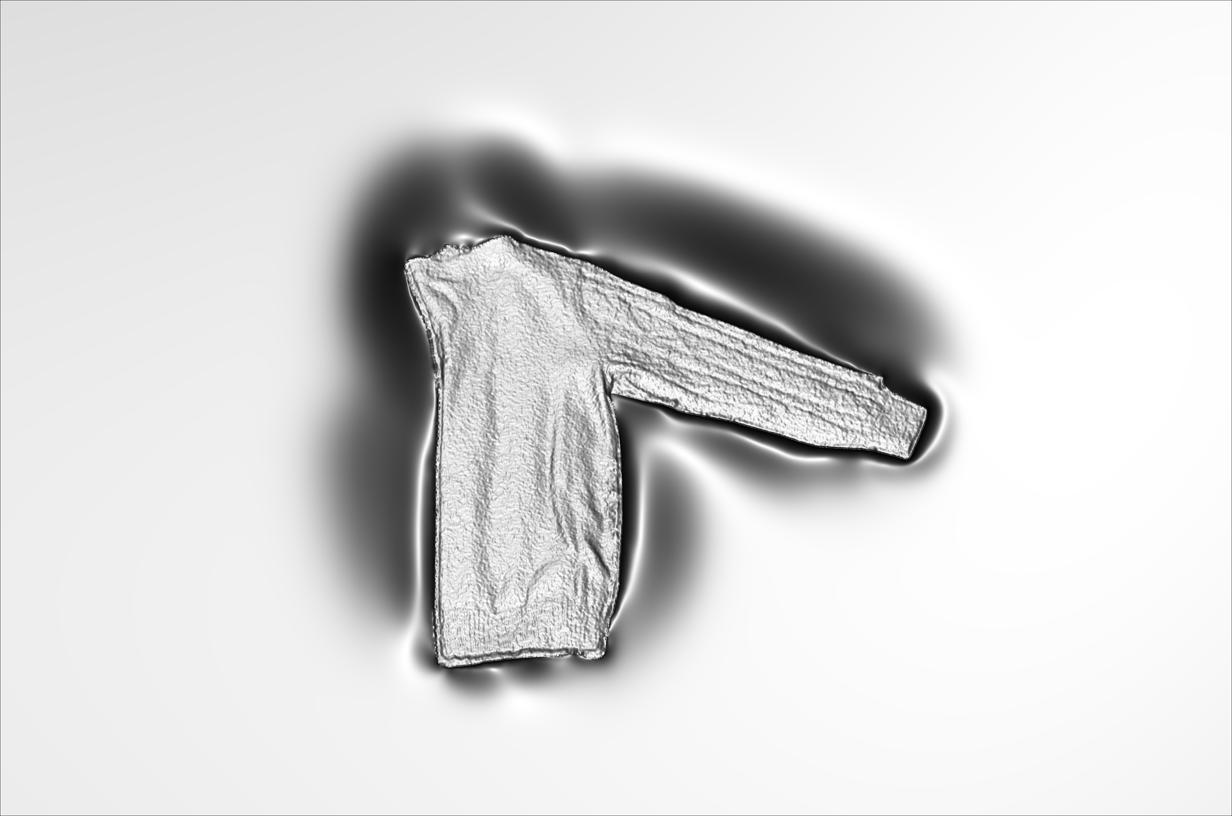}\tabularnewline
\end{tabular}

\caption{Examples of range images computed at different zoom settings.\label{fig:Examples-of-range}}
\end{figure}

The underlying objective of the stereo matching algorithm is to locate
for each pixel in one image of a stereo pair, the corresponding location
on the other image of the pair. The correspondence problem is solved
by constructing a displacement field (also termed parallax or disparity
map) that maps points in the left image to the corresponding location
on the right image. These displacement fields are expressed in terms
of two disparity maps for storing horizontal and vertical displacements
mapping pixels in the left image to the corresponding location in
the right image. Computed disparities can then be used to reconstruct
highly detailed point clouds and/or range images. Range image preview
examples can be depicted in Figure \ref{fig:Examples-of-range}. It
should be noted that point clouds and range images are not included
in the database as the file size of each stereo-pair sample is roughly
in the order of 1GB; however, source code to recover the 3D geometry
from the disparity maps is included in the database.

\section{Database File Description and Organisation}

\begin{figure}
\centering\includegraphics[width=10cm]{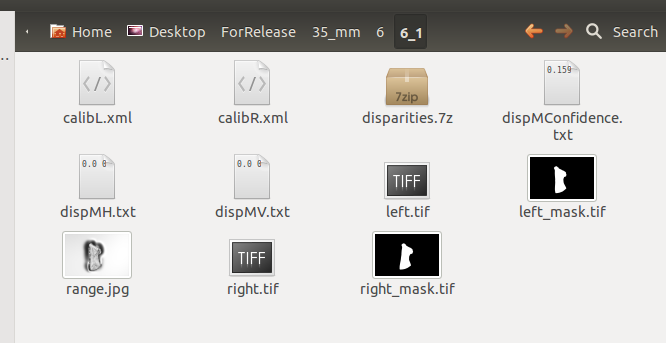}

\caption{Example of the file organisation of the stereo database.\label{fig:Example-of-the}}
\end{figure}

The database (\url{https://sites.google.com/site/ugstereodatabase/})
is firstly divided according to the captured garments. These are organised
and stored in folders using a numeric index from 1 to 16. In each
of these folders, garment pose configurations are organised in folders
which follow the the following file format:\textbf{ XX\_S}; where
\textbf{XX} denotes the garment class and the folder number where
the image is stored and \textbf{S}, the garment pose configurations.
\textbf{S} can take the following classification indices which correspond
to how the garment was captured:
\begin{itemize}
\item 0 - Cloth is spread on the table (Figure \ref{fig:Garment-states}(a)).
\item 1 - Cloth is half-way folded (Figure \ref{fig:Garment-states}(b)).
\item 2 - Cloth is completely folded (Figure \ref{fig:Garment-states}(c)).
\item 3 - Cloth is wrinkled (Figure \ref{fig:Garment-states}(d)).
\item 4 - The robot is holding the cloth in the air and close to the table
(Figure \ref{fig:Garment-states}(e)).
\end{itemize}
Within the above folders, the following is stored (it can also be
depicted in Figure \ref{fig:Example-of-the}):
\begin{itemize}
\item Stereo-pair images (left and right camera images) are stored as 16Mpixel
colour TIFF image files (4928 x 3264 x 24 BPP).
\item Annotated image masks for the stereo-pair are stored as black and
white TIFF files, i.e. (4928 x 3264 x 8 BPP).
\item Horizontal (\emph{dispMH}) and vertical (\emph{dispMV}) disparity
maps and a confidence matching map (\emph{dispMConfidence}) are stored
as text files, in ASCII format, as matrices of 4928 by 3264 floating
point values. These maps are compressed as 7zip format.
\item A JPEG compressed preview of the garment range image.
\end{itemize}

Camera calibration parameters are stored as XML files for each of
the captured garments. Calibration files are saved as \emph{calL.xml}
and \emph{calR.xml} for the left and right cameras, respectively,
as showed in Figure \ref{fig:Example-of-the}. These XML files can
be easily read using OpenCV I/O XML functions. The companion source
code provides an example on how to load these calibration files. Calibration
parameters in each file include:
\begin{itemize}
\item Camera matrix, $K$, as a 3 by 3 matrix that stores the focal point
and principal point in pixels.
\item Distortion coefficients, $D$, as a 1 by 4 vector. The Glasgow stereo
matcher and stereo reconstruction does not use this information; however,
this coefficients are included for completeness.
\item Projection matrix, $P$, as a 3 by 4 matrix. This matrix is defined
for the left (Equation \ref{eq:1}) and right (Equation \ref{eq:2})
cameras as follows:
\begin{equation}
P_{L}=K_{L}\left[\mathrm{I}|0\right]\label{eq:1}
\end{equation}
\begin{equation}
P_{R}=K_{R}[R|t]\label{eq:2}
\end{equation}

where $\mathrm{I}$ is a 3 by 3 identity matrix, $R$ and $t$, the
rotation and translation matrices that transforms the right camera
reference frame into the left camera reference frame. $P_{L}$ and
$P_{R}$ are used to recover the 3D structure of the captured scene.

\item Fundamental matrix, $F$, as a 3 by 3 matrix that relates corresponding
points between the stereo-pair. The same numeric matrix is defined
in both files.
\end{itemize}

\section*{Acknowledgements}

We would like to thank the European Community\textquoteright{}s Seventh
Framework Programme (FP7/2007-2013) to support this research work
under grant agreement no 288553, CloPeMa.

\section*{References}

\bibliographystyle{elsarticle-harv}
\bibliography{references}

\end{document}